\documentstyle[aaai]{article}

\begin{document}

\title{Metrics for evaluating dialogue strategies \\
 in a spoken language system}

\author{Morena Danieli and  Elisabetta Gerbino\\
CSELT - Centro Studi e Laboratori Telecomunicazioni \\
Via G. Reiss Romoli 274 - 10148 Torino, Italy \\
E-Mail: danieli and gerbino@cselt.stet.it}

\date{}
\maketitle
\begin{abstract}
\begin{quote}
In this paper, we describe a set of metrics for the evaluation of
different dialogue management strategies in
an implemented real-time spoken language system. The set of metrics we propose
tries to offer
useful insights in evaluating how particular choices in the dialogue
management can affect the overall quality of the man-machine
dialogue. The evaluation
makes use of established metrics: the transaction
success, the contextual appropriateness of 
system answers, the calculation of normal and correction
turns in a dialogue. We also define a new metric, the implicit recovery,
which allows
to measure the ability of a dialogue manager to deal with errors by
different levels of analysis. We report evaluation data from several 
experiments, and we compare two different approaches to dialogue
repair strategies using the set of metrics we argue for.
\end{quote}
\end{abstract}

\section{Introduction}

A dialogue module which is part of a complex natural language system (for
example, of a speech understanding system providing information) may be
evaluated according to different viewpoints, the more important of which are:

(1) its ability to drive the user to find the required information;

(2) the overall quality of the dialogic interaction;

(3) its capacity to maintain an acceptable level of interaction with the user,
also when other modules have partial or total breakdowns.

The first point may be measured in terms of the success of the dialogic
transaction, but the second 
and the third points are rather matters of subjective evaluation. 
This dichotomy
is reflected in the state of the art of dialogue evaluation methods.
While there is a set of objective metrics which can be used to measure the 
performance of a dialogue system \cite{Hirschman90}, only in the last few
years an effort has been done to define metrics which express 
the subjective evaluation of dialogue systems.

By taking into account the shortcomings of 
the recent work done on subjective evaluation \cite{Simpson93}, 
\cite{Hirschman93}, this paper has the goal 
of arguing for a set of metrics which can be fruitfully used to compare 
the behaviour of different dialogue strategies within speech systems.
In particular, we introduce a new metric
(implicit recovery) which captures the ability of the dialogue manager to
recover from partial or total failure of previous levels of analysis.
Other metrics we used 
(i.e. contextual appropriateness,
turn correction ratio, transaction success) derive from the 
set of evaluation metrics defined within the Sundial Esprit 
project \cite{danieli92}: in
what follows they are discussed only to point out possible
difference in their interpretation. 

We will show the application of these
metrics to evaluate data coming from two trials carried out on 
the  spoken man-machine dialogue system developed at CSELT for
the Italian language. The application domain of the system is the 
Italian railway time-table; the system allows to access a remote database
using the telephone and to get information about the train times and
services. During the experimentation two different approaches to
dialogue management were tested and evaluated. This methodology 
of evaluation allows to compare the different effects
of the two approaches on the system behaviour.

\section{Metrics and Methods of Evaluation}

\subsection{Implicit Recovery}

In evaluating a dialogue strategy for a spoken language system,
attention should be paid to capture its capacity
to deal with situations in which errors occur. In particular,
we expect that a well conceived dialogue system should be
able to repair from both partial and total failure by the previous 
levels of analysis. That feature may
be considered according to different points of view: on one hand the
dialogue manager should be able to filter the parser output and to interpret
it starting from contextual knowledge. On the other hand, it should embody
explicit strategies to recover from understanding or recognition errors.
While the latter ability may be evaluated in terms of the number
of correction turns undertaken by system and user (see below), we define 
the {\em implicit recovery} (IR) as a measure of the former ability.

The IR is the measure of the dialogue module capacity to regain 
utterances which are partially failed at recognition or understanding
levels. When the linguistic processor
performs a robust partial parsing, the dialogue module
may receive either a correctly representation of the utterance conceptual
content, or only partial results. Of course, in spite of robustness of the
parser, complete failure in understandings may also occur. 
Utterances which have been partially misunderstood may have
insertions of concepts which are not present in the original 
utterance, deletions of some concepts or substitutions of the value
of a concept
with another one. The dialogue module should be able to deal with 
this kind of errors by interpreting them within the dialogue context.

In order to measure the IR, we need a semantic representation formalism 
that allows to calculate the percentage of
correctly understood concepts. In evaluating the experimental data (see 
the fourth paragraph), we 
used the conceptual accuracy metric (ConA) at the syntactico-semantic level. 
\footnote{This is a metric which 
uses at the understanding level the word accuracy formula, 
expressed in terms of insertion, deletion and substitution of concepts, as
in \cite{baggiacrim94}}

To apply the IR metric, an expert examines the dialogue logfiles and
for each user's utterance, he checks if the semantic representation of
the utterance meaning given by the parser is correct or not.
No IR occurs if the utterance has been correctly understood 
or completely failed.
Otherwise the expert sees if the error has been
recovered,i.e. an appropriate answer is given by the system.
\footnote{For the definition of contextual appropriateness see the
  next paragraph.}
 This case is marked to be an IR. 
The IR final result is the percentage between the number of cases where the
dialogue manager was able to correct the conceptual errors and the number
of sentences which presents conceptual errors.

\begin{figure}[htb]
\begin{small}
\begin{tabular}{l} \hline \\

{\bf U1}: I want to go from Roma to Milano in the morning. \\

$<$arrival-city=MILANO, departure-time=MORNING$>$ \\

{\bf S1}: Sorry, where do you want to leave from? \\

{\bf U2}: From Roma. \\

$<$departure-city=ROMA, cost-of-ticket?$>$ \\

{\bf S2}: Do you want to go from Roma to Milano leaving \\
in the morning?  \\ \\
\hline 
\end{tabular}
\end{small}
\caption{\label{implrec}Example of Implicit Recovery}
\end{figure}
 
Figure ~\ref{implrec} shows an example of dialogue interaction where
two IR occur. In the first dialogue turn, the user's utterance
contains all the concepts the system needs to retrieve the desired
information, but the recognition (or parsing) level fails to represent
the departure city. The dialogue takes into account the correctly
understood concepts and asks for the concept which was lost.
In the second turn, the recognition level inserted some words in the
best decoded sequence that the parser interprets as a request of the
cost of ticket. But since for the dialogue strategy that concept is
not relevant in the current context, the system does not consider it
and asks the user to confirm only the correct concepts
it has been able to collect. In similar cases we would say that the IR
percentage is 100\%.

\subsection{Other Metrics}

The contextual appropriateness is a measure of the degree of contextual
coherence of the system answers. The concept of {\em contextual
appropriateness} (CA) is taken from the Grice's conversational maxims
\cite{LC} and it has been used within the Sundial
project to evaluate the appropriateness of system utterances in their
dialogue context. We have restricted the definition of contextual
appropriateness proposed in \cite{Simpson93} to obtain a three-valued measure:
appropriate, inappropriate and ambiguous.

According to this restricted interpretation, we say that a system utterance is
appropriate (AP) when it provides the user with the information he required,
when it asks him to give additional constraints which are
essential to interpret his request or when it introduces (or continues)
a repair strategy. A system utterance is inappropriate (IA) when it
supplies the user with wrong information or when it fails to interpret
the speaker's utterance in the correct context. Finally, a system utterance
is ambiguous (AM) when it violates the Gricean maxims of quantity and manner,
i.e. it is over (or under) informative, it is obscure and it is not
orderly and  brief. 

During the implementation of the dialogue systems we saw that it
was useful to measure the ratio of those
turns which are concerned with anomalous behaviour from both the
user and the system to all turns in a dialogue; we named this measure
{\em turn correction ratio} (TCR). 
The TCR is calculated adding the results of the application of two
submetrics: the turn correction by the system (STC) and the turn correction
by the user (UTC). The STC concerned those dialogue turns where
the system introduces a recovery strategy and tells the user to
repeat or rephrase his sentence. The UTC occurs when the user 
detects or corrects an error,  repeats or rephrases an utterance.

All the turns which are neither STC nor UTC are considered normal turns: 
by following the classification proposed in \cite{Hirschman93}, we consider 
normal turns of the system the appropriate directives, such as the
introductory message, the appropriate diagnostic
messages and the correct answers. The normal turns of the user are the
utterances used to request information (both first and continuation
utterances),
and the answers to appropriate system directives. 

Finally, we used the concept of {\em transaction success} (TS) to measure
the success of the system in providing the speakers  with the information they
required, when such information is available in its database.

\subsection{Methodology}

The system configuration permitted to store all the data collected in the
tests:
the speech material, the semantic representation of the sentences (parser 
output), the dialogue logfile (user/system interactions) and some timing 
(recognition time, parser time and dialogue time). 
All the speech material had to be manually transcribed; the 
dialogue corpus evaluation was performed by two experts on the dialogue 
logfiles as in \cite{Goodine92}.

Subjects' global impressions
were collected by asking subjects to complete a questionnaire. Comments
on questionnaire are in \cite{Ciar93}.

\section{Description of the Experimental Set-up}

Two different trials were carried out along three months on an  
integrated spoken man-machine dialogue system which
allows the access to a remote DB using the telephone. This prototype 
was partially developed under the Sundial Esprit Project.
The application domain consists of the Italian train time-table information.
The first trial was carried out in March 1993 and the second one in May 1993.
For the first trial, twenty subjects were recruited among 
people who have never used a computerized 
telephone service before. 
Those subjects were paid for testing the system;
ten out of them were female and ten were male; the average age of the
subjects was 37.

For the second trial, fifteen people were recruited among CSELT staff:
eleven out of them were male, four were female.
The average age of the male subjects was 35, that of the female
subjects was 30.

The subjects came to CSELT laboratories and received a single page of printed
directions which contained a brief explanation of the service capabilities
and some instructions (e.g.: ``Please, speak after the tone'').
All the subjects carried out the test being alone in an isolated room. 
During the dialogue with the system they had
to get information about train time-table and related services
(sleeping-cars, restaurant, rates and extra-fares, reservation
and so on).

To precisely determine if the task has been
solved, predefined pictorial scenarios were used. Each scenario specified the 
departure and arrival city names, chosen among the set of 100 cities of the 
railway DB in use, and the train attributes to be collected during the
dialogue, while the user was free to specify the departure time.

In both trials each subject had to play at least 4 scenarios;
the corpus
of dialogues collected in the tests are shown in Table ~\ref{corpus}. 
For each trial, the total number of dialogues, the number of
continuous speech utterances, and the average number of words per
utterance are reported.

\begin{table}[htb]
\begin{center}
\begin{tabular}{|l|c|c|c|c|} \hline \hline
Trial & No. of & No. of & No. of & Avg. words \\ 
& Subj. & Dial. & Utt. & per Utt.\\ \hline
1st & 20 & 85 & 678 & 4.8 \\
2nd & 15 & 63 & 464 & 4.2 \\
\hline \hline
\end{tabular}
\end{center}
\caption{\label{corpus}Dialogues corpus characteristics}
\end{table}

\subsection{Overview of the System Architecture}
\label{system}

The system is composed by the following modules:
 the acoustical front-end (AFE), the linguistic processor (LP),
 the dialogue manager and message generator (DM), and
 the text-to-speech synthesizer. 
The acoustical front-end and the synthesizer are interconnected to the PBX
through a telephone interface, while the dialogue manager is connected to a 
Computer Information System to obtain the information on Italian
train time-tables. The system is nearly real time.
For a complete description of the system see \cite{Clem93}.

The AFE performs feature extraction and acoustic-phonetic decoding;
both DDHMM and CDHMM are supported and the vocabulary size is about 800 words
\cite{Rico}. In these experiments we used a preliminary version of the
recognition module which used always DDHMM 
with Forward decoding algorithm and made no use of linguistic models.

The LP starts from the AFE output, best-decoded sequence,
and performs a multi-step robust partial parsing.
In this strategy, partial solutions are accepted according to
the linguistic knowledge \cite{ICASSP93}. 
At the end of the parsing stage a deep semantic
representation for the user utterance is sent to the DM.

The DM models the user-system interaction and contextually interprets
sentences using a prediction mechanism \cite{Bet93}.
The DM also implements many recovery strategies
in order to allow completion of the task, even in case of lower level errors.
Finally, it generates the answer sent to the 
text-to-speech synthesizer, which contains specific prosodic rules
oriented to the dialogue~\cite{Quazza93}.

During the trials, two different versions of the DM 
(D1 and D2) were
tested by all subjects; each carried out two telephone calls with one version 
and two calls with the other one. 

\begin{figure}[htb]
\begin{small}
\begin{tabular}{l} \hline \\
{\bf S1}: Hello, This is Train Enquiry Service. Please speak \\
after the tone. Which information do you need?  \\

{\bf U1}: I want to go from Torino to Milano. \\

    $<$departure-city=TRENTO, arrival-city=MILANO$>$\\

{\bf S2}: Do you want to go from Trento to Milano? Yes \\
or No?\\

{\bf U2}: No. \\

    $<$confirmation=NO$>$\\

{\bf S3}: Do you want to leave from Trento? Yes or No? \\

{\bf U3}: No. \\

    $<$confirmation=NO$>$ \\

{\bf S4}: Where do you want to leave from? \\

{\bf U4}: I want to leave from Torino. \\

    $<$departure-city=TORINO$>$ \\

{\bf S5}: Do you want to leave from Torino? Yes or No? \\

{\bf U5}: Yes. \\

     $<$confirmation=YES$>$ \\

{\bf S6}: Do you want to go to Milano? Yes or No? \\

{\bf U6}: Yes. \\

    $<$confirmation=YES$>$ \\

{\bf S7}: At which time do you want to leave? \\

{\bf U7}: I want to travel in the evening. \\

    $<$departure-time=EVENING$>$ \\

{\bf S8}: Do you want to leave between 6 and 9 p.m.? \\
Yes or No? \\

{\bf U8}: Yes. \\

    $<$confirmation=YES$>$ \\

{\bf S9}: There is a train from Torino Porta Nuova to Milano \\
Centrale. It leaves at 7.10 p.m. and arrives at 8.55 \\ 
p.m.. Do you want more detailed information about  \\
it? \\

{\bf U9}: Is that an intercity train and how much does the \\
ticket cost? \\ \\
\hline 
\end{tabular}
\end{small}
\caption{\label{D1ex}Example of D1 dialogue interaction}
\end{figure}

\begin{figure}[htb]
\begin{small}
\begin{tabular}{l} \hline \\
{\bf S1}: Hello, This is Train Enquiry Service. Please speak \\
after the tone. Which information do you need? \\

{\bf U1}: I want to travel from Milano to Roma. \\

    $<$departure-city=MERANO, arrival-city=ROMA$>$ \\

{\bf S2}: At which time do you want to leave from Merano \\
to Roma?  \\

{\bf U2}: No, I want to leave from Milano in the evening. \\

    $<$confirmation=NO, departure-city=MILANO, \\
    departure-time=EVENING$>$ \\

{\bf S3}: Do you want to leave from Milano between 6 and \\
11 p.m.? Please answer Yes or No? \\

{\bf U3}: Yes. \\

    $<$confirmation=YES$>$ \\

{\bf S4}: Intercity 243 leaves from Milano Centrale at 20 \\
past 8 p.m.; it arrives at Roma Termini 6 o'clock a.m.. \\
Do you want another train information? \\

{\bf U4}: No \\

    $<$confirmation=NO$>$ \\

{\bf S5}: Good bye. \\ \\
\hline 
\end{tabular}
\end{small}
\caption{\label{D2ex}Example of D2 dialogue interaction}
\end{figure}
 
The D1 version was totally developed at CSELT and it is currently 
used in the integrated system.
The D2 version was developed under the Sundial Esprit Project.

Figure ~\ref{D1ex} shows an example of dialogue interaction carried out
with the D1 version, while Figure ~\ref{D2ex} shows an example of 
dialogue interaction carried out with the D2 version.
As we can observe from the dialogue examples, the more salient differences
between the two DM versions concerned 
the kind of information supplied and the repair strategies. 

The D1 version provides the opportunity to ask 
information about train times and services, but it does not leave the 
user free to make requests about more than one train route in a single 
telephone call. On the contrary, the D2 version allows the request of 
information about 
many train routes in a single telephone call, but it does not provide 
the caller with information about the train services.
Moreover the D1 confirmation and repair strategies are specifically
implemented to deal with
possible speech errors. Actually the D1 (see Fig.~\ref{D1ex})
implements an explicit concept confirmation strategy forcing the
user to answer with isolated words (yes on no); this strategy is
robust and safe even if it increases the number of turns spent in
confirmations and consequently the dialogue time.
Besides, the D1 strategy makes use of more robust speech interaction
modalities such as isolated word and spelling. 

The D2 system implements discourse strategies which are more confident in the 
capabilities of the acoustic and linguistic processors. Actually, D2
is able to support three confirmation strategies: confirmation alone
for a bunch of concepts; confirmation for a bunch of concepts plus
initiative; confirmation concept by concept, and
then initiative. However the dialogue manager is not able to switch
autonomously from one strategy to another when it detects troubles
with the communication. When an error occurs, D2 decides to enter a
special mode: after three requests for repetition, the last one using the
spelling modality, the system advises the user to contact a human operator. 

An example of D2 multiple confirmation plus initiative strategy is shown in
Figure ~\ref{D2ex}. There the U1 utterance is
misunderstood at the recognition level. The dialogue module implicitly asks
for confirmation of departure and arrival cities by asking for the
desired departure time (see S2). In U2 the subject denies the departure
city proposed by the system, reconfirms that  he wants to leave from Milano
and gives the system the departure time. System utterance  S3 shows
that D2 considers the arrival city as implicitly confirmed and carries on
the interaction by focusing on the new acquired concepts. During the
testing of this system we chose to run it with a confirmation
strategy which did not forced the subjects to have recourse to
isolated word recognition.

\section{Evaluation Results}

The dialogue corpus collected in the trials was analysed according to 
the whole set of evaluation metrics. At the recognition and
understanding levels, users' utterances were evaluated by considering
the standard measurements: respectively, the Word Accuracy (WA) 
calculated on the best decoded
sequence against the transcribed uttered sentence, and the Sentence
Understanding (SU).
In the first trial the results were: 52.1\% of WA and 50.9\% of SU.
In the second trial the results were: 60.2\% of WA and 59.1\% of SU.
\footnote{Recent recognition results are available  in
\cite{giachin95}. Now the system obtains 82.6\% of WA by using
linguistic models at the recognition level.}
As regards these data, we did not 
distinguish between the two different DM versions because
they are related to the two common system modules (AFE and LP).

As regards the dialogue level, we calculated contextual appropriateness (CA),
explicit recovery (ER)
and implicit recovery (IR). Moreover we distinguished between the two DM 
versions,
in order to study the capability of these metrics to point out the 
differences between various dialogue strategies.
Table ~\ref{cart} shows the results obtained in the trials for the
contextual appropriateness. The first column reports the percentage of
the appropriate sentences uttered by the system; the second column
reports the percentage of the inappropriate sentences and the third
column shows the percentage of ambiguous utterances. As we can see,
both the dialogue systems are seldom ambiguous; that say us that the
generation modules of both the systems are good.

\begin{table}[htb]
\begin{center}
\begin{tabular}{|l|c|c|c|} \hline \hline
\multicolumn{1}{|l|}{Trial} &
\multicolumn{3}{c|}{CA} \\
 & AP & IA & AM  \\ \hline
1st D1 & 77.6\% & 20.6\% & 1.8\% \\
1st D2 & 49.2\% & 50.3\% & 0.5\%  \\ \hline
2nd D1 & 79.1\% & 19.3\% & 1.6\%  \\
2nd D2 & 56.5\% & 43.5\% & 0.0\%  \\
\hline \hline
\end{tabular}
\end{center}
\caption{\label{cart}Contextual Appropriateness Results}
\end{table}

We deem that the contextual appropriateness metric is useful to evaluate 
the quality of the dialogic interaction and to address the issue of 
co-operation in human computer dialogue.
We can expect that in a ideally perfect speech system, where no recognition 
and understanding errors occur, 
the CA should measure properly the DM capability to correctly interpret
user's utterances. Starting from the same percentage of correctly
understood utterances (respectively 50.9\% and 59.1\% in the two
trials), 
D1 and D2 get very different CA scores. 
In particular, AP results reflect the
greater robustness of the D1 version when it faces off difficulties at
the recognition or understanding level.

Another data which stands out is the growth of the percentage of
AP when the users are good conversationalists with the computer, as
the subjects participating to the second trial were. The value of AP
increases more for the D2 dialogue system: that means that the
dialogue strategies it implements are more sensitive to the
performances of the previous levels of analysis.

\begin{table}[htb]
\begin{center}
\begin{tabular}{|l|c|c|c|} \hline \hline
\multicolumn{1}{|l|}{Trial} &
\multicolumn{2}{c|}{ER} &
\multicolumn{1}{c|}{IR} \\
 & UTC & STC &  \\ \hline
1st D1 &  24.8\% & 31.8\% & 17.0\% \\
1st D2 &  67.9\% & 65.6\% & 10.8\% \\ \hline
2nd D1 &  25.6\% & 22.5\% & 17.0\% \\
2nd D2 &  45.0\% & 49.2\% & 10.7\% \\
\hline \hline
\end{tabular}
\end{center}
\caption{\label{er}Recovery Results}
\end{table}

Table ~\ref{er} shows the results obtained in the trials for the
metrics which measure the recoveries from errors implemented both by
the systems and by the subjects. If we consider the ER data, we can
read in the first column the percentage of correction turns done by
the users (UTC), while the percentage of correction turns by the
systems (STC) is
reported in the second column. The experiments highlight that if
a dialogue strategy is not robust enough to deal with errors by the lower 
levels (see D2 data), the number of turns
spent by both user and system in repairing from errors grows up.
When users are more co-operative, as staff subjects were,
the percentage of STC and UTC decreases.

The third column of Table ~\ref{er} reports the percentage of turns in
which the dialogue systems implicitly recovered from errors by
recognition and understanding. Those data show that the capacity of
implicit recovery of D1 and 
D2 does not vary from naive to expert users:
actually, IR is a measure
of a dialogue system ability and it does not depend upon
the degree of users' co-operation.
Since D1 and D2 make use of different degrees of predictive contextual
knowledge, we expected a difference in their IR performance and that is 
shown by the data. The different performance is also due to 
the fact that D1 applies its predictive knowledge in more and more focused
interpretation contexts as far as the dialogue goes on. For example,
the recourse to the request for a single concept, see Figure 
~\ref{D1ex} turns S4 and S7, allows using focused predictive
knowledge.  

On the contrary, the interpretative focus of D2 is always wider,
so that it cannot exploit the advantages of very constrained 
contextual interpretation, which seems to be useful in this kind of
application of discourse analysis. Let us consider the
use of implicit confirmation strategy showed in Figure ~\ref{D2ex},
turn S2. There the reply to system enquiry by the subject may contain
a great deal of information, for instance the negation of what the
system said along with the introduction of new concepts (see turn U2).
In this case, the use of very constrained predictive knowledge is
hardly possible.

Table ~\ref{sys} shows the whole system performance: the percentage of 
TS, the average number of turns per dialogue, the average 
dialogue time, and the TCR.

\begin{table}[htb]
\begin{center}
\begin{tabular}{|l|c|c|c|c|} \hline \hline
Trial & TS & Avg. No. & Avg. Dial. &  TCR \\ 
& & of Turns & Time & \\ \hline
1st D1 & 77.6\% & 20 & 5'15" & 10.0\%  \\
1st D2 & 51.0\% & 11 & 3'20" & 27.0\% \\   \hline
2nd D1  & 96.6\% & 21 & 5'09" & 9.5\%  \\
2nd D2  & 83.3\% & 11 & 2'59" & 15.0\%  \\
\hline \hline
\end{tabular}
\end{center}
\caption{\label{sys}Whole System Performances}
\end{table}

The TS is always good, but it increases as the users are more friendly
or as much as the acoustic and linguistic processors have better
performance.

Finally, we notice that the number of turns and the dialogue time are
higher with D1:
this difference is due to the fact that D1 allows the request of many 
information about train services and it does not close the interaction if
there are difficulties in recognition or understanding.

\section{Conclusions}

The results of these experiments are encouraging as regards the effectiveness 
of the metrics we used. We have argued that it is important to capture
the ability of a dialogue system to reduce the consequences of 
recognition and understanding errors. 

The necessity of many specific
metrics is due to the fact that 
various dialogue strategy aspects have to be evaluated. Actually,
at least three aspects have to be measured: the
dialogue system ability to drive the user to find the desired
information is captured by measuring the transaction success along with
the average number of turns in the dialogue, while the quality of the
man-machine interaction is measured by the metric of contextual
appropriateness.
Finally,  the dialogue system robustness is evaluated by
measuring its ability to perform both implicit and explicit
recoveries when the lower levels of the system fail.
This set of metrics  also enables the dialogue system designer to
verify  the success of alternative  strategies. According to us, 
in this field another important
research topic should be the definition of methods for evaluating the
system effectiveness and friendliness from the user's point of view.

Before concluding this paper, we would like to thank Sheyla Militello
for her help during the experimentation activity.

\end{document}